\documentclass{article}
\usepackage{spconf,amsmath,graphicx}
\usepackage{threeparttable}
\usepackage{dcolumn}
\usepackage{booktabs}
\usepackage{amsmath,graphicx}
\usepackage{url}
\usepackage{cite}
\usepackage{psfrag}
\usepackage{amsfonts,amssymb}
\usepackage{amsbsy}
\usepackage{graphicx}
\usepackage{epstopdf}
\usepackage{array}
\usepackage{multirow}
\usepackage{bm}
\usepackage{subfigure}
\usepackage{verbatim}
\usepackage{colortbl}
\usepackage{stfloats}
\usepackage{algorithm}
\usepackage{algorithmic}
\usepackage{setspace}

\usepackage{bigstrut}
\usepackage{amsmath}
\usepackage{mathtools}

\def\L{{\cal L}}

\title{Explaining the Predictions of Any Image Classifier via Decision Trees}
\name{Sheng Shi$^{\dagger}$ \qquad Xinfeng Zhang$^{\star}$ \qquad Wei Fan$^{\dagger}$}
\address{$^{\dagger}$ AI Laboratory, Lenovo Research, Beijing 100094, China \\
$^{\star}$ University of Chinese Academy of Sciences, Beijing 100049, China}

\begin{document}
%\ninept
%
\maketitle
\begin{abstract}
Despite outstanding contribution to the significant progress of Artificial Intelligence (AI), deep learning models remain mostly black boxes, which are extremely weak in explainability of the reasoning process and prediction results. Explainability is not only a gateway between AI and society but also a powerful tool to detect flaws in the model and biases in the data. Local Interpretable Model-agnostic Explanation (LIME) is a recent approach that uses an interpretable model to form a local explanation for the individual prediction result. The current implementation of LIME adopts the linear regression as its interpretable function. However, being so restricted and usually over-simplifying the relationships, linear models fail in situations where nonlinear associations and interactions exist among features and prediction results. This paper implements a decision Tree-based LIME approach, which uses a decision tree model to form an interpretable representation that is locally faithful to the original model. Tree-LIME approach can capture nonlinear interactions among features in the data and creates plausible explanations. Various experiments show that the Tree-LIME explanation of multiple black-box models can achieve more reliable performance in terms of understandability, fidelity, and efficiency.
\end{abstract}
\section{Introduction}
\label{sec:intro}
In recent years, the fast-growing computing power, enormous consumer and commercial data, and emerging advanced machine learning algorithms jointly stimulate the prosperous of AI \cite{AI01}\cite{AI02}, which has gone from a science-fiction dream to a critical part of our daily life. Compared to traditional machine learning methods, deep learning has achieved superior performance in perception tasks such as object detection and classification. However, because of the nested non-linear structure, deep learning models usually remain black boxes that are particularly weak in the explainability of the reasoning process and prediction results. In many real-world mission-critical applications, transparency of deep learning models and explainability of the model outputs are essential and necessary in their real deployment process.

Explanable AI is not only a gateway between AI and society but also a powerful tool to detect flaws in the model and biases in the data. The development of techniques on explainability and transparency of deep learning models has recently received much attention in the research community \cite{One01}\cite{One02}\cite{One03}\cite{One04}. The relevant research roughly falls into two categories: global explainability and local explainability. Global explainability aims at making the reasoning process wholly transparent and comprehensive \cite{global03}\cite{global04}, while local explainability focuses on extracting input regions that are highly sensitive to the network output to provide explanations for each decision \cite{local01}\cite{local02}\cite{local03}\cite{local04}.

An effective way to achieve explainability is to use a light-weight function family to create interpretable models. Local interpretable model-agnostic explanations (LIME) identify an interpretable model over the human-interpretable representation that is locally faithful to the original model \cite{local01}. The current implementation of LIME in \cite{local01} adopts the linear regression as its interpretable function, which represents the prediction as a linear combination of a few selected features to make the prediction process transparent (Linear-LIME). However, being so restricted and usually over-simplifying the relationships, linear regression models fail in some situations where non-linear associations and interactions exist among features and prediction results %\cite{ref01}.

The LIME framework in \cite{local01} supports the exploration of a variety of explanation families, such as linear regression and decision trees. In this paper, we implement a Decision Tree-based Local Interpretable Model-agnostic Explanation (Tree-LIME). The decision tree structure creates good explanations as the data ends up in distinct groups that are often easy to understand. Moreover, the tree structure can capture interactions between features in the data. We perform various experiments on explaining two black-box models, the random-forest classifier and Google's pre-trained Inception neural network\cite{Inception}. The results show that Tree-LIME explanations achieve more reliable performance than Linear-LIME in terms of understandability, fidelity, and efficiency.

\section{Interpretable models}
\label{sec:format}
Using a subset of algorithms from a light-weight function family to create interpretable models is an effective way to achieve interpretability. In this section, we analyze two representative interpretable models - the linear regression model and the decision tree model. Table~\ref{tab:1} shows the properties of two interpretable models. The linear regression displays the prediction as a linear combination of features, while the decision tree represents the reasoning process in a hierarchical structure, which is suitable for capturing the nonlinear association between features and predictions. The monotonicity constraint shown in both models is necessary to ensure the consistency between a feature and the target outcome. Moreover, the decision tree model can automatically capture the diverse interactions between features to predict the target outcome, applicable to both classification and regression tasks.   %In this paper, we mainly focus on the image classifier model.

\begin{figure}
\begin{minipage}{1\linewidth}
  \centering
  \centerline{\includegraphics[width=1.0\textwidth]{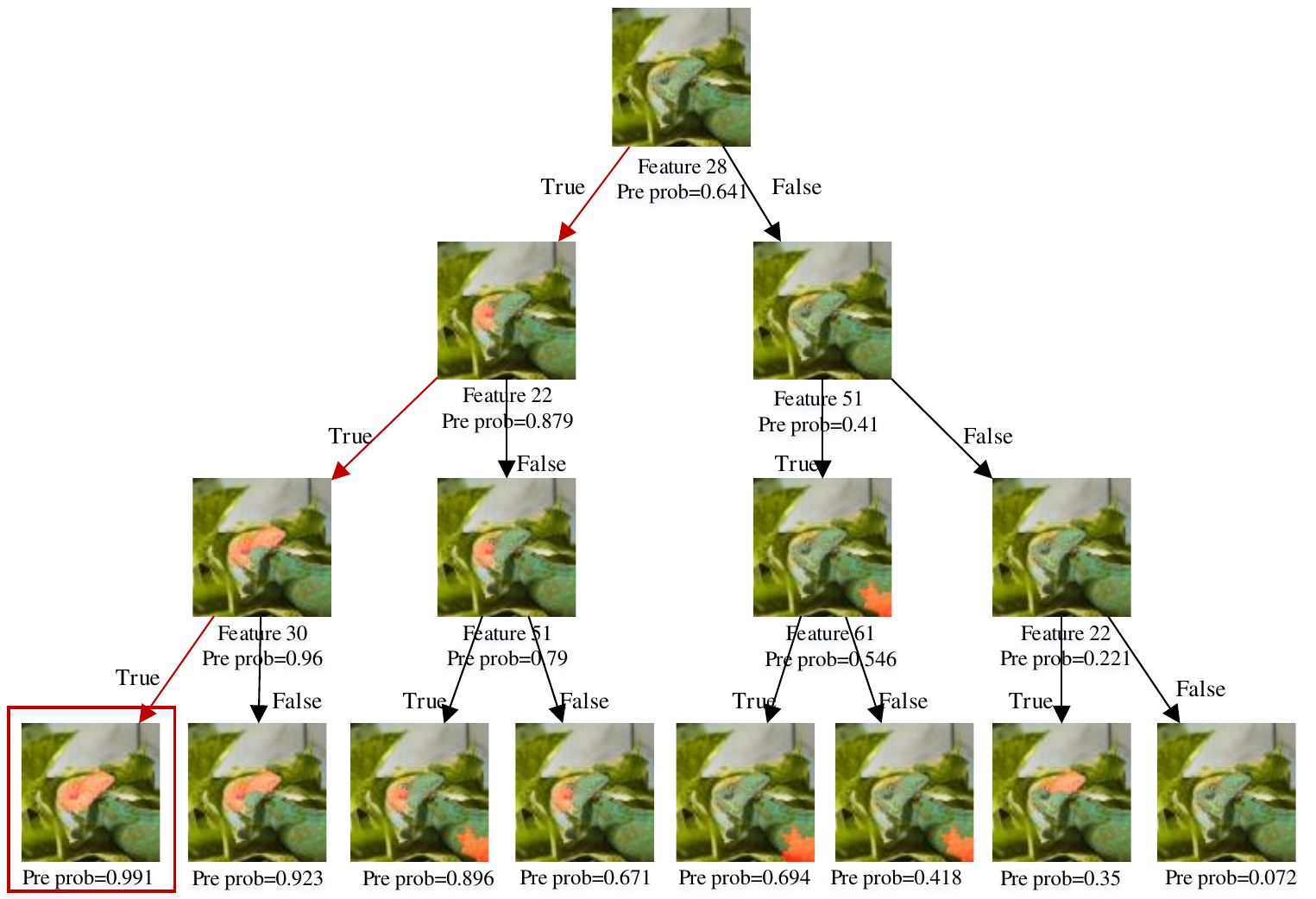}}
\end{minipage}
\caption{\small{A simple regression decision tree explains the image classification results given by Google's Inception neural network. The top $1$ output class label is $African$ $chanmeleon$ with the prediction probability $(p=0.9935)$.}}
\vspace{-10pt}
\label{fig:1}
\end{figure}

\begin{table*}
  \centering\footnotesize
  \caption{\small{The properties of linear regression model and decision tree model}}
    \begin{tabular}{|c|c|c|c|c|}
    %\begin{tabular}{|c|c|c|c|}
    \hline
    Models          & Linearity  & Monotonicity  & Feature Interaction  & Task   \bigstrut\\
    \hline
    Line regression & Yes     & Yes       & No           & Regression
    \bigstrut\\
    \hline
    Decision tree  & No    & Some   & Yes    & Classification, Regression\bigstrut\\
    \hline
    \end{tabular}%
  \label{tab:1}%
\end{table*}
Depending on the different criteria, various algorithms are capable of constructing a decision tree. The CART \cite{ref02} is the most popular algorithm which can handle both classification and regression tasks. In this paper, we mainly construct regression decision trees to explain the prediction probability of the image classifier. Figure~\ref{fig:1} illustrates a simple regression tree to explain image classification prediction made by Google's Inception neural network. The predicted top $1$ class label is $African$ $chanmeleon$ $(p=0.9935)$. The highlighted superpixels give intuition as to why the model would choose that class. The decision tree shows that if feature $28$, $22$, and $30$ exist, then the prediction probability is $0.991$, which is the mean value of the instances $y$ in this node. Moreover, The importance of the three features is $0.7164$, $0.0709$, and $0.0259$, showing the contribution of the three features in improving the variance.

%\end{figure}

\section{The Tree-LIME Approach}
\label{sec:pagestyle}
In this section, we first present the interpretable image representations and the decision tree-based LIME approach. Then we discuss the characteristics of Tree-LIME.

%\subsection{Interpretable Image Representations}
\subsection{Explanation System of Tree-LIME}

\begin{figure}
\begin{minipage}{0.32\linewidth}
  %\centering
  \centerline{\includegraphics[width=1.0\textwidth]{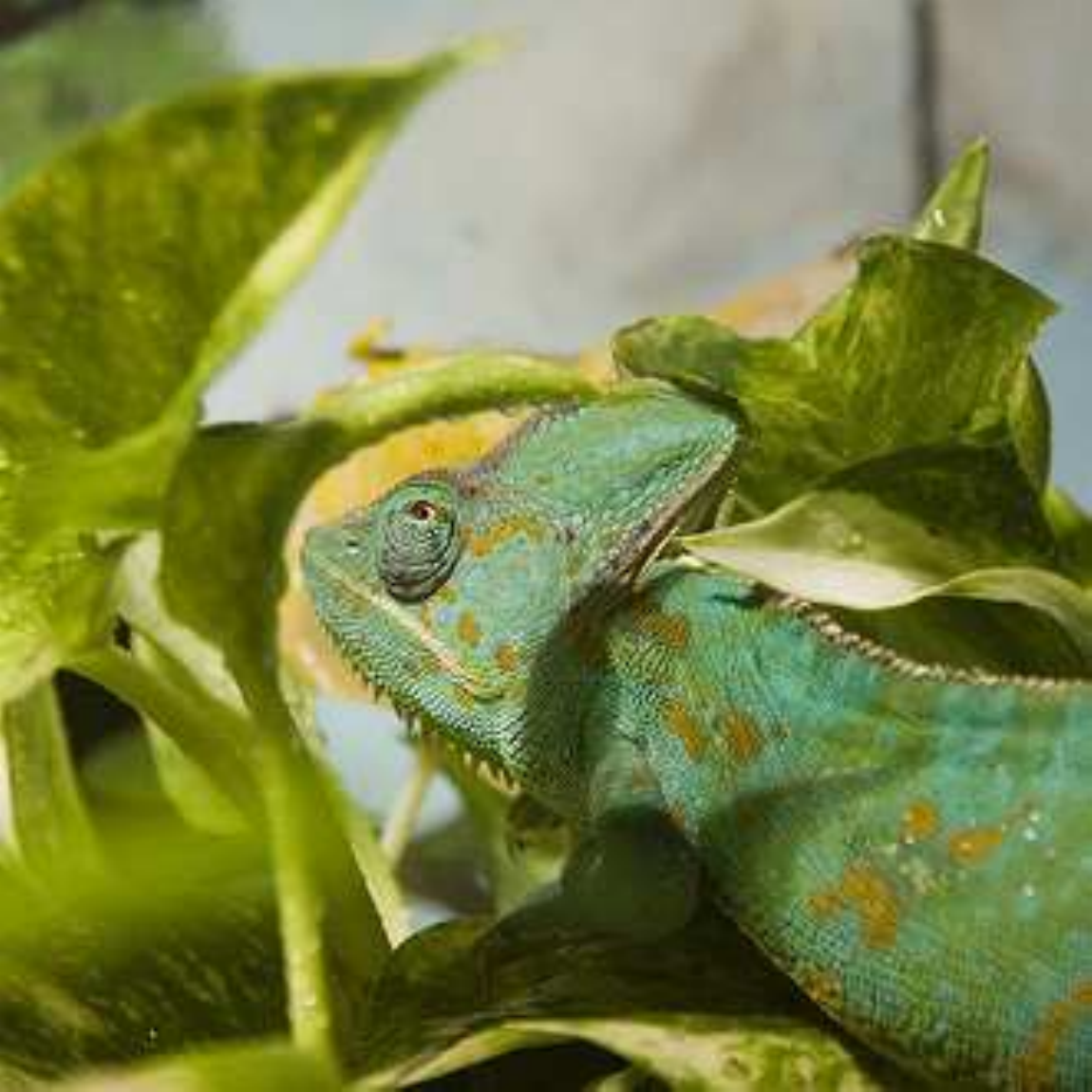}}
  \centerline{\scriptsize{(a) Pixel-based image}}
  %\centerline{}
\end{minipage}
%\wfill
\begin{minipage}{0.32\linewidth}
  %\centering
  \centerline{\includegraphics[width=1.0\textwidth]{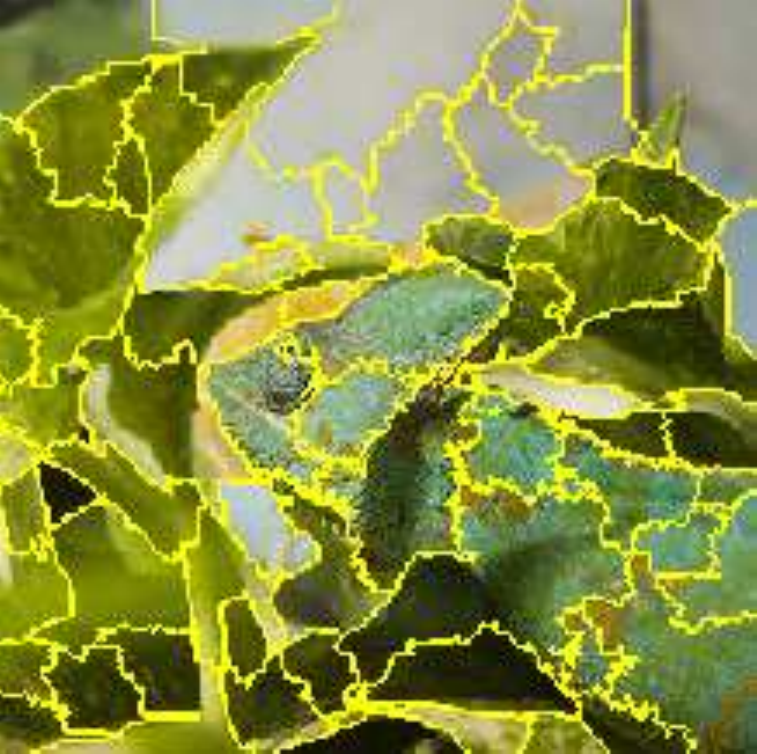}}
  \centerline{\scriptsize{(b) Superpixel image}}
  %\centerline{}
\end{minipage}
\begin{minipage}{0.32\linewidth}
  %\centering
  \centerline{\includegraphics[width=1.0\textwidth]{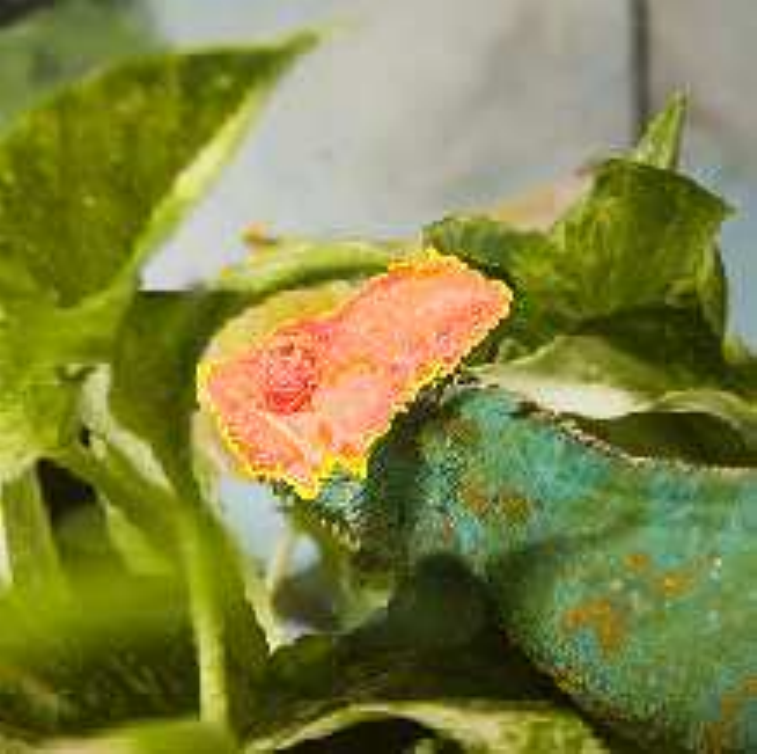}}
  \centerline{\scriptsize{(c) Superpixel-based explanation}}
  %\centerline{}
\end{minipage}
\caption{\small{Pixel-based image and superpixel image}}
\label{fig:3}
\end{figure}

Considering the poor interpretability and high computational complexity of the pixel-based image representation, we adopt a superpixel based explanation system. Each superpixel, as the basic processing unit, is a group of connected pixels with similar colors or gray levels. Figure~\ref{fig:3} shows the pixel-based image, superpixel image, and superpixel-based explanation. The interpretable representation of an image $x\in{\mathbb{R}^d} $ consisting of $d$ pixels and $d'$ superpixels is a binary vector $x'\in{\{{0,1}\}^{d'}}$ where $1$ indicates the presence of original superpixel and $0$ indicates absence of original superpixel.

%Moreover, we use the classicial felzenszwalb's graph-based segmentation method \cite{fel} to get superpixel image as shown in Figure~\ref{fig:3}.

We denote the original image classification model being explained as $f$, the interpretable decision tree model as $g$, and the locality fidelity loss as $L(f,g,\pi_x)$, which is calculated by the locally weighted square loss:

\begin{equation}
\L(f,g,\pi_x)=\sum_{z,z'\in Z}e^{(-D(x,z)^2/{\sigma}^2)}(f(z)-g (z'))^2.\\
\end{equation}

The database $Z$ is composed of perturbed samples $z'\in{\{{0,1}\}^{d'}}$ which are sampled around $x'$ by drawing nonzero elements at random. Given a perturbed sample $z'$, we recover the sample in the original representation $z\in {\mathbb{R}^d}$ and get $f(z)$. Moreover, $\pi_x(z)=exp(-D(x,z)^2/{\sigma}^2)$ where distance function $D$ is the $L_2$ distance of image $x$ and $z$ is used to capture locality.

We denote the decision tree explanation produced by Tree-LIME as below:
\begin{equation}
\xi(x)=argmin\quad{L(f,g,\pi_x)+ dep(g)}.
\end{equation}
The depth of decision tree $dep(g)$ is a measure of model complexity. A smaller depth indicates a stronger understandability of model $g$. In order to ensure both local fidelity and understandability, formula (2) minimizes locality-fidelity loss $L(f,g,\pi_x)$ while holding $dep(g)$ low enough. Algorithm 1 shows a simplified workflow diagram of Tree-LIME. Firstly, Tree-LIME gets the superpixel image by using a standard segmentation method. Then the database $Z$ is constructed by running multiple iterations of the perturbed sampling operation. Finally, within the allowable range of prediction error, Tree-LIME gets the minimum depth decision tree by using the CART method.

\begin{algorithm}[h]
\caption{\small{Decision Tree based Local interpretable model-agnostic explanation (Tree-LIME)}}
\label{alg::conjugateGradient}
\begin{algorithmic}[1]
\footnotesize
\REQUIRE
Classifier $f$,
Number of samples $N$,
Instance $x$,
Max depth of tree $d$
\ENSURE
time and prediction error of Tree-LIME
\STATE get superpixel image $x'$ by segment method
\STATE initial $Z=\{\}$
\FOR{$i=1$; $i<N$; $i++$}
\STATE get $z'$ by sampling around $x'$
\STATE get $f(x')$ by classifier $f$
\STATE get $z$ by recovering $z'$
\STATE $Z \leftarrow Z \cup (z'_i,f(z_i),\pi_x(z_i))$
\ENDFOR
%\STATE initial $dep=1$;
\FOR{$j=1$; $j<d$ and $error<\delta$; $j++$}
%\STATE get decision tree $g$ by CART algorithm;
\STATE get decision tree $g=CART(Z, maxdepth=j)$
\STATE $error=\|f(x)-g(x')\|$
\ENDFOR
\STATE output decision tree $g$, time and prediction error
\end{algorithmic}
\end{algorithm}

\subsection{Characteristics of Tree-LIME}
\begin{figure}
\begin{minipage}{1\linewidth}
  \centering
  \centerline{\includegraphics[width=0.5\textwidth]{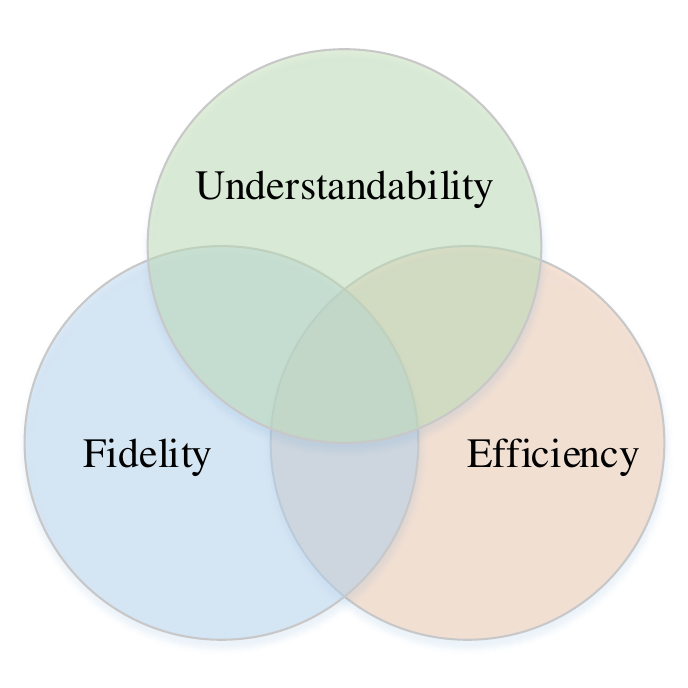}}
  %\centerline{\scriptsize{(b) Decision Tree Expanation}}
  %\centerline{Decision Tree Explanation}
\end{minipage}
\caption{\small{The three goals of Interpretability}}
  %\vspace{-10pt}
\label{fig:2}
\end{figure}

Despite the fact that the amount of research in explainable AI is growing actively, there is no universal consensus on the exact definition of interpretability and its measurement criterion \cite{ref01}. Ruping first noted that interpretability is composed of three goals - accuracy, understandability, and efficiency \cite{ref03}. We argue that fidelity is a better description than accuracy since accuracy is easily confused with the performance evaluation criteria of the original black box model. These three goals are inextricably intertwined and competing with each other, as shown in Figure~\ref{fig:2}. An explainable model with good interpretability should be faithful to the data and the original model, understandable to the observer and graspable in a short time so that the end-users can make wise decisions.

%This section describes the local interpretable model-agnostic explanation enabled with decision tree model (TLIME).
Tree-LIME has many appealing characteristics, such as interpretable, local fidelity, and model-agnostic. It provides a qualitative understanding of features and predictions. It is challenging, if not impossible, to be utterly faithful to the black box model on a global scale. Tree-LIME takes a feasible approach by approximating it in the vicinity of an instance being predicted. Besides, Tree-LIME, as a model-agnostic interpretation, shows excellent flexibility and capability of explaining any underlying machine learning model.

\section{Experimental Results}
\label{sec:typestyle}

In this section, Tree-LIME and Linear-LIME \cite{local01} explain the predictions of RandomForeset Classifier and Google's pre-trained Inception neural network. We compare the experimental results of the two algorithms in terms of understandability, fidelity, and efficiency. %All experiments were carried out on the system that is equiped with 8 Intel Xeon E7-8830 2.13GHz CPU total of 64 cores and 1TB memory. we present numerical experimental results

\subsection{RandomForeset Classifier on MNIST database}
The MNIST database is one of the most common databases used for image classification. It consists of $7\times{10^5}$ small $28\times28$ grayscale images of handwritten digits. In this experiment, the image data is split into $70\%$ as the training set and $30\%$ as the test set. Table~\ref{tab:2} shows the performance of the random forest classifier. For instance $x$, the predicted top $1$ classe is $Seven$ (p=1.0). Figure~\ref{fig:6} shows the decision tree explanations by Tree-LIME.
%For instance $x$, the prediction probability matrix is $f(x)=[0,1.0,0,0,0,0,0,0,0,0,0]$.Some representative images are shown in Figure~\ref{fig:5}.
\begin{table}[h]
  \centering\footnotesize
  \caption{\small{The performance of randomforeset classifier on MNIST database}}
    \begin{tabular}{|c|c|c|c|c|}
    \hline
              & precision  & recall  & f1-score  & support   \bigstrut\\
    \hline
    weighted avg & 0.95 & 0.95 & 0.95  & 21000
    \bigstrut\\
    \hline
    \end{tabular}%
   \label{tab:2}%
\end{table}

%\subsubsection{Understandability}
Comparing with Linear-LIME, which can only provide a one-shot explanation, the decision tree structure by Tree-LIME provides a more intuitive explanation. Figure~\ref{fig:6} shows that if feature $0$ and feature $3$ exist, then the prediction probability is $1.0$. Moreover, the tree structure can capture the interaction between features in the data. The importance of feature $0$ and $3$ is $0.9416$ and $0.0402$, respectively, which tells us the feature $0$ makes a significant contribution to predicting the outcome. The prediction error is calculated to measure local fidelity. The prediction error of Tree-LIME is $0.0$, showing better fidelity than Linear-LIME with an error of $0.0529$.

Efficiency is highly related to the time necessary for a user to grasp the explanation. The runtime of CART in Tree-LIME is $0.0020s$, which is faster than that of K-LASS0 in Linear-LIME \cite{local01} - $0.0080s$.%Moreover, the experiment results by running experiments 5000 times are shown in Table~\ref{tab:3}.
%TLIME creates good explanations as the data ends up in distinct groups that are often easy to understand.

\begin{figure}
\begin{minipage}{1\linewidth}
  \centering
  \centerline{\includegraphics[width=0.9\textwidth]{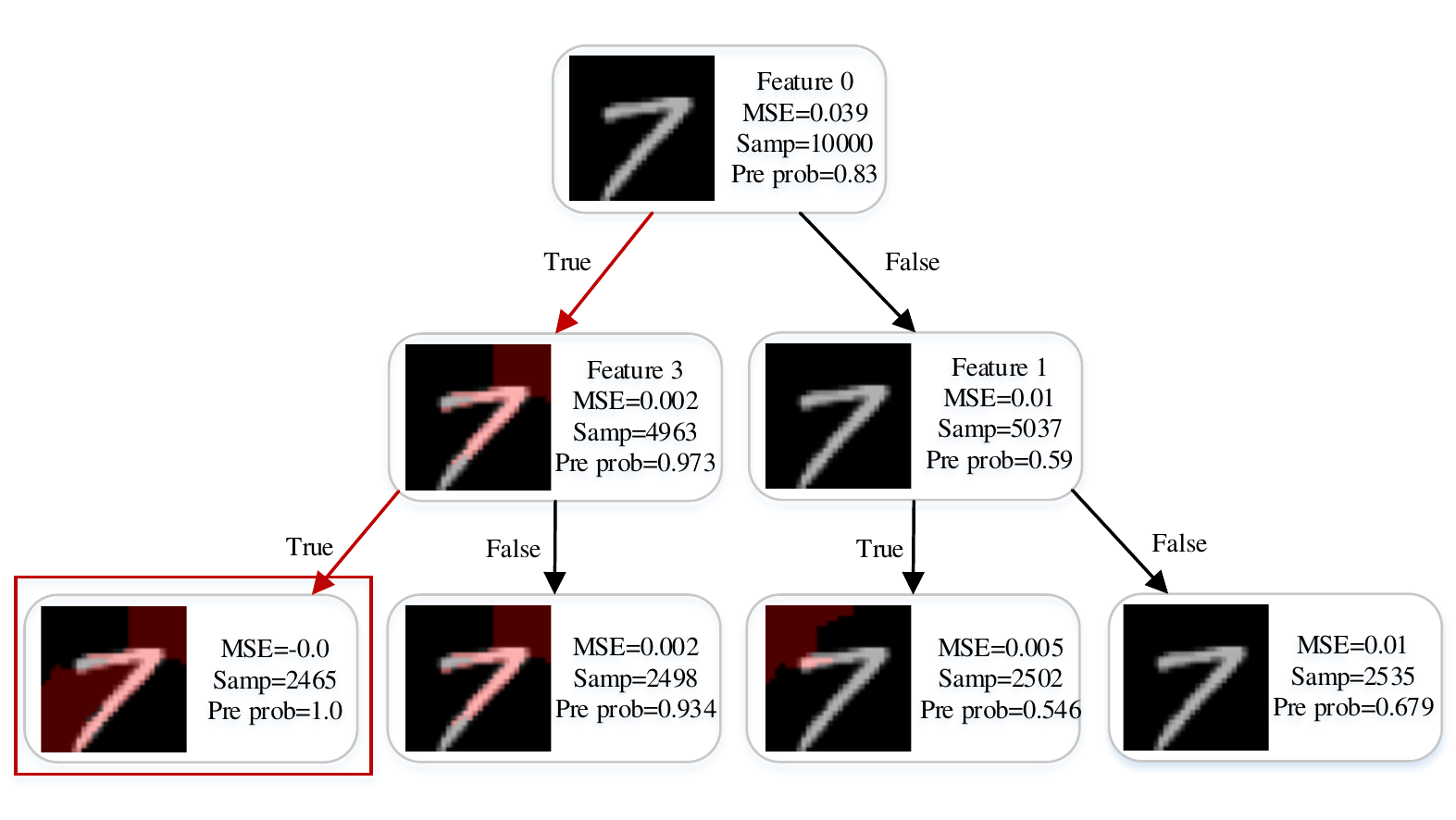}}
\end{minipage}
\small
\caption{\small{Explaining an image classification prediction made by randomforeset classifier. The top $1$ class predicted is $seven$ $(p=1.0)$}}
%\vspace{-10pt}
\label{fig:6}
\end{figure}

\begin{table}
  \centering\footnotesize
  \caption{\small{The prediction probability and prediction errors of  Linear-LIME and Tree-LIME on Google's Inception neural network}}
    \begin{tabular}{|c|c|c|c|c|}
    %\begin{tabular}{|c|c|c|c|}
    \hline
     & Inception prob & pred prob  & pred error \bigstrut\\
    \hline
    Linear-LIME  & \multirow{2}{*}{$p(lamp)=0.4785$}& 0.6264 & 0.1479 \bigstrut\\
    %\hline
    Tree-LIME & & 0.4309 & 0.0476 \bigstrut\\
    \hline
    Linear-LIME  &\multirow{2}{*}{$p(couch)=0.4203$} & 0.6814 & 0.2611 \bigstrut\\
    %\hline
    Tree-LIME & & 0.4682 & 0.0479 \bigstrut\\
    \hline
    Linear-LIME  &\multirow{2}{*}{$p(pillow)=0.0227$} & 0.0168 & 0.0059 \bigstrut\\
    %\hline
    Tree-LIME & & 0.0191 & 0.0036 \bigstrut\\
    \hline
    \end{tabular}
    %\vspace{-50pt}
  \label{tab:3}
\end{table}

\begin{table}
  \centering\footnotesize
  \caption{\small{The Time of Linear-LIME and Tree-LIME on Inception neural network}}
    \begin{tabular}{|c|c|c|c|}
    %\begin{tabular}{|c|c|c|c|}
    \hline
      & $table$ $lamp$ & $studio$ $couch$  & $pillow$ \bigstrut\\
    \hline
    %\hline
     K-LASS0 in Linear-LIME  & 0.0289s & 0.0150s & 0.0120s\bigstrut\\
    \hline
    CART in Tree-LIME & 0.0060s & 0.0030s &0.0090s\bigstrut\\
    \hline
    \end{tabular}
  \label{tab:4}
\end{table}

\subsection{Google's Inception neural network on Image-net database}
\begin{figure}
\begin{minipage}{0.24\linewidth}
  \centerline{\includegraphics[width=1\textwidth]{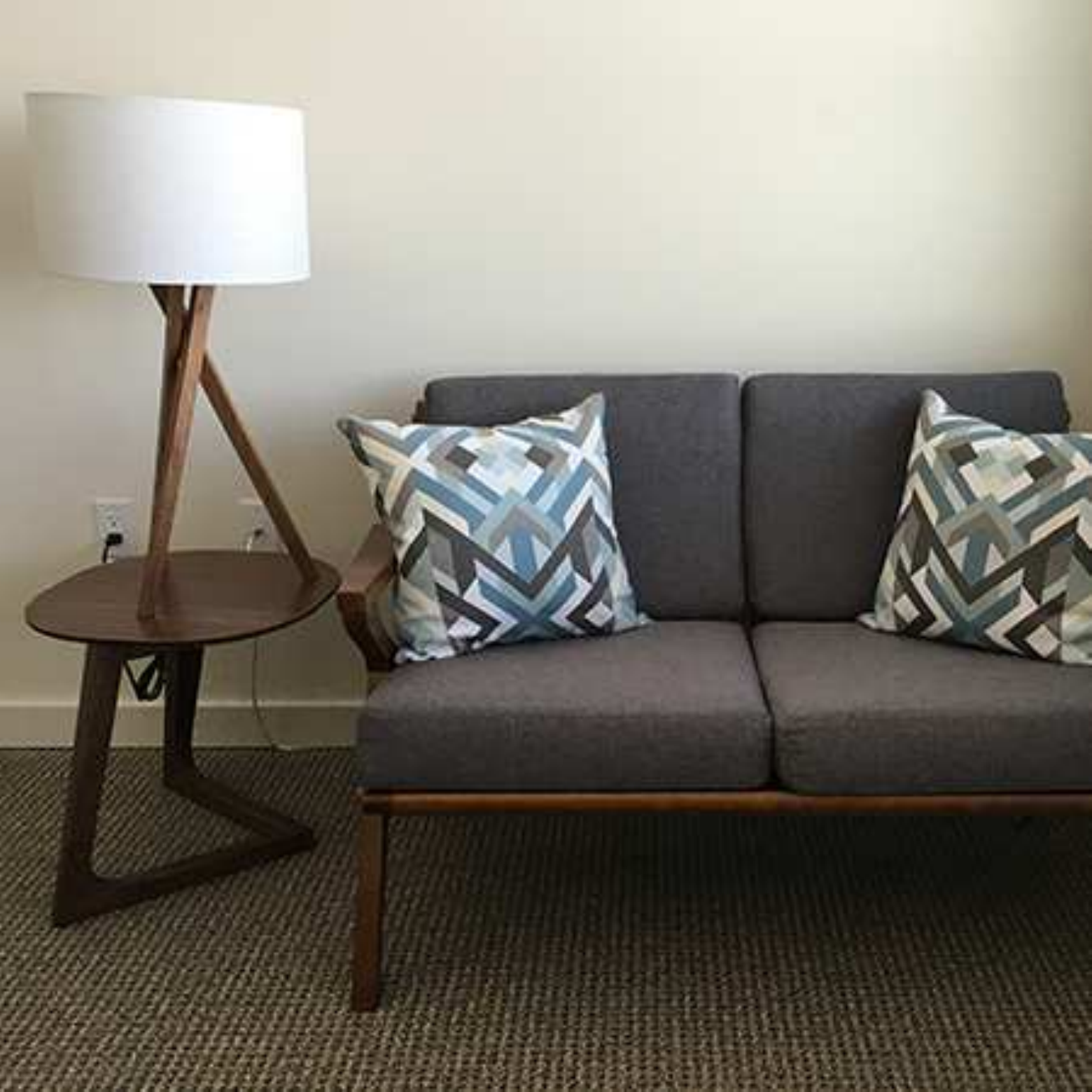}}
  \centerline{\scriptsize{(a)}}
  %\centerline{}
\end{minipage}
\hfill
\begin{minipage}{0.24\linewidth}
  \centerline{\includegraphics[width=1\textwidth]{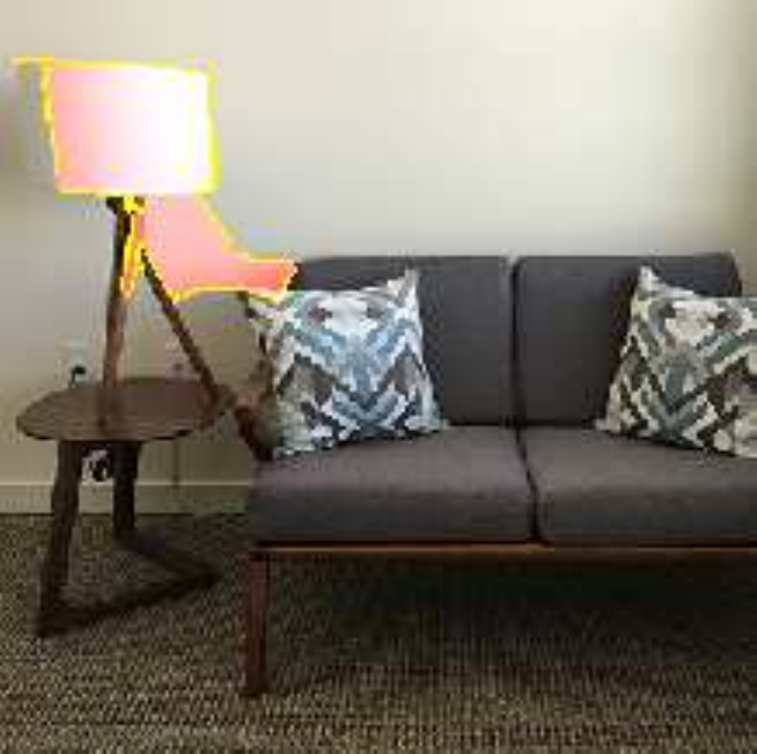}}
  \centerline{\scriptsize{(b)}}
  %\centerline{}
\end{minipage}
\hfill
\begin{minipage}{0.24\linewidth}
  \centerline{\includegraphics[width=1\textwidth]{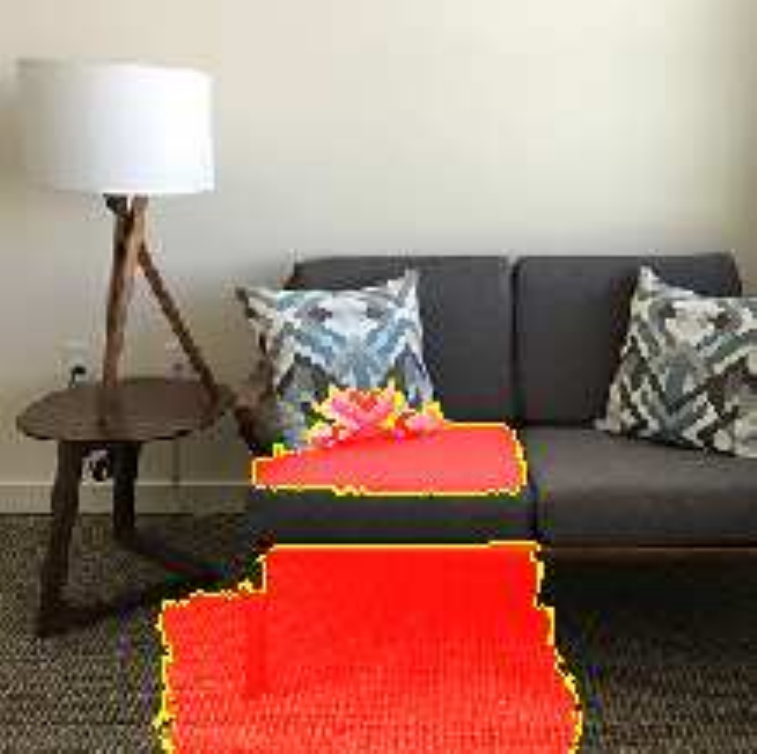}}
  \centerline{\scriptsize{(c)}}
  %\centerline{}
\end{minipage}
\hfill
\begin{minipage}{0.24\linewidth}
  \centerline{\includegraphics[width=1\textwidth]{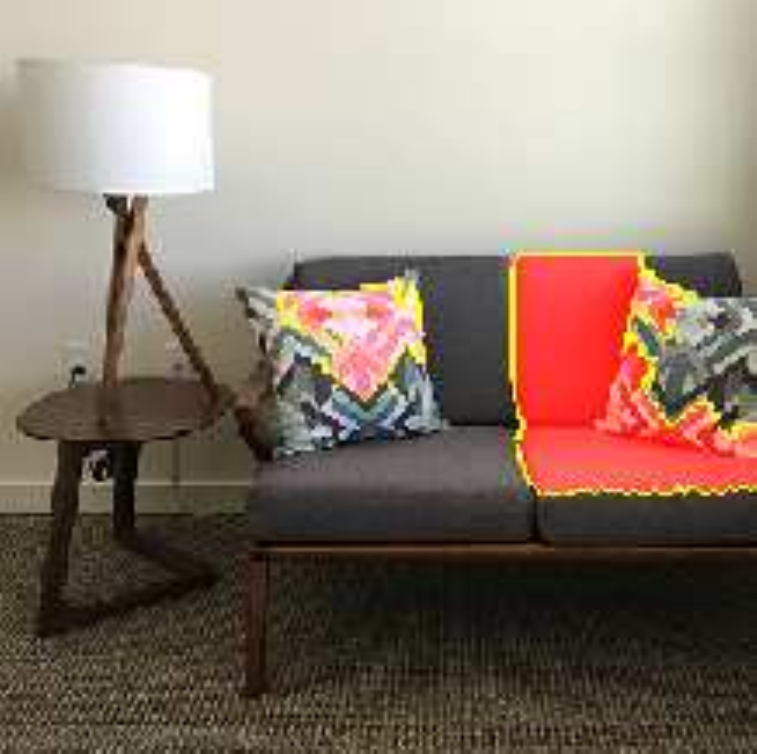}}
  \centerline{\scriptsize{(d)}}
  %\centerline{}
\end{minipage}
\vfill
\caption{\small{Explaining an image classification prediction made by Google's Inception neural network using Tree-LIME. The top $3$ classes predicted are $table$ $lamp (p=0.4785)$, $studio$ $couch (p=0.4203)$, $pillow (p=0.0227)$}: (a) Original image. (b) Explaining $table$ $lamp$. (c) Explaining $studio$ $couch$. (d) Explaining $pillow$.}
\vspace{-10pt}
\label{fig:5}
\end{figure}

We explain the prediction of Google's pre-trained Inception neural network on the image shown in Figure~\ref{fig:5}(a). Figures~\ref{fig:5}(b), \ref{fig:5}(c) and \ref{fig:5}(d) show the superpixels explainations for the top $3$ predicted classes: $table$ $lamp (p=0.4785)$, $studio$ $couch (p=0.4203)$ and $pillow (p=0.0227)$ respectively. Due to space constraints, Figure~\ref{fig:6} shows the the explainations of $table$ $lamp$ and $studio$ $couch$ by Tree-LIME. The prediction provides reasonable insight into what the neural network picks upon for each of the classes. This kind of explanation enhances trust in the classifier. Moreover, Table~\ref{tab:3} lists the prediction errors of Tree-LIME and Linear-LIME. The prediction error shows that Tree-LIME has a better fidelity than Linear-LIME. Table~\ref{tab:4} lists the runtime of CART in Tree-LIME and runtime of K-LASSO in Linear-LIME.

\begin{figure}
\begin{minipage}{0.49\linewidth}
  \centerline{\includegraphics[width=1\textwidth]{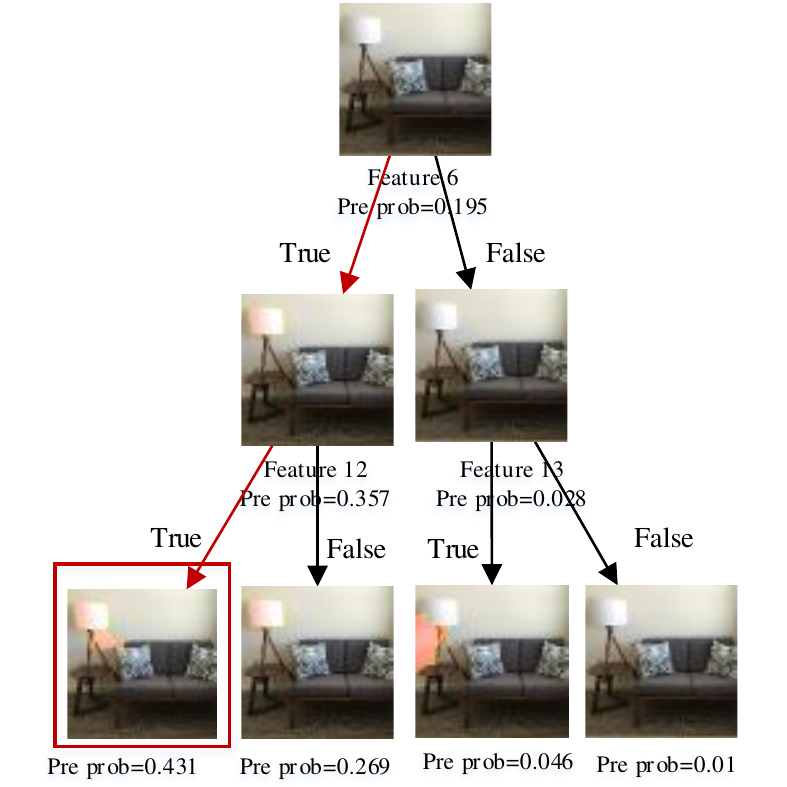}}
  \centerline{\scriptsize{(a) Explaining $table$ $lamp$}}
  %\centerline{}
\end{minipage}
\hfill
\begin{minipage}{0.49\linewidth}
  \centerline{\includegraphics[width=1\textwidth]{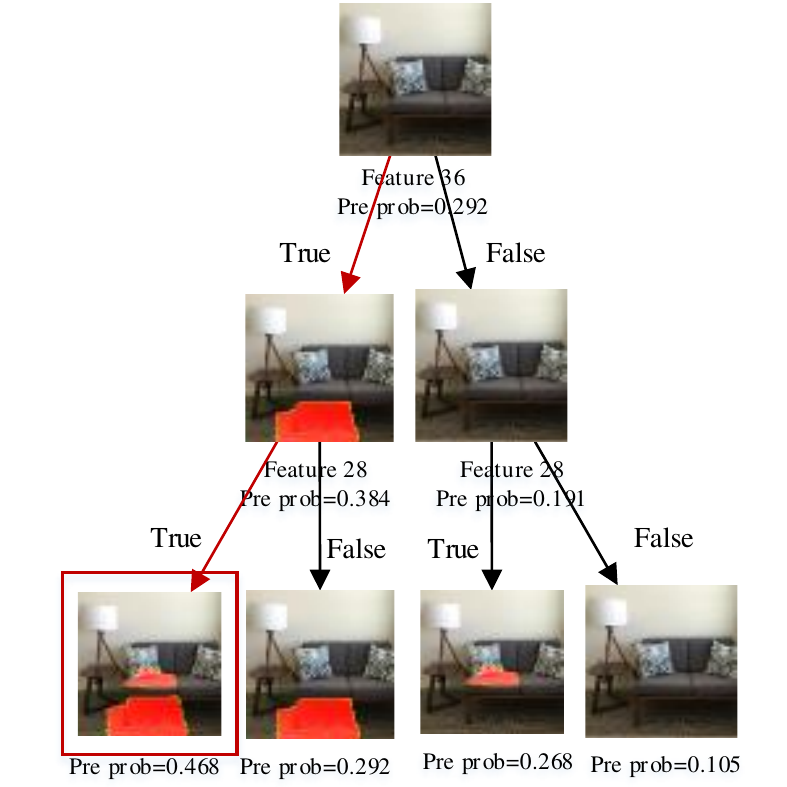}}
  \centerline{\scriptsize{(b) Explaining $studio$ $couch$}}
  %\centerline{}
\end{minipage}
%\hfill
%\begin{minipage}{0.32\linewidth}
%  \centerline{\includegraphics[width=1\textwidth]{figures/process/office_simple_721.pdf}}
%  \centerline{\scriptsize{(c)}}
  %\centerline{}
%\end{minipage}
%\small
\caption{\small{Explaining an image classification prediction made by Google's Inception neural network using Tree-LIME.}}
\vspace{-10pt}
\label{fig:6}
\end{figure}

%Figure~\ref{fig:7} show the superpixels explanations for the top $3$ predicted classes.
%\begin{figure}

We can conclude from the above results that under less time, Tree-LIME not only has a better understandability but also has a higher fidelity than Linear-LIME.

\section{Conclusion}
The LIME framework in \cite{local01} supports the exploration of a variety of explanation families. The current implementation of LIME adopts the linear regression as its interpretable function (Linear-LIME). In this paper, we implement a decision tree-based local interpretable model-agnostic explanation (Tree-LIME). The goal of Tree-LIME is to construct an interpretable decision tree model over the interpretable representation that is locally faithful to the oringal classifier. We compare Tree-LIME and Linear-LIME in explaining the predictions of RandomForeset Classifier and Google's pre-trained Inception neural network. Experimental results have shown that Tree-LIME exhibits a better understandability and higher fidelity than Linear-LIME using less process time, which covers the ingredients of an ideal explainable AI model - understandability, fidelity, and efficiency.
%Moreover, the tree structure can capture interactions between features in the data.

\newpage
\bibliographystyle{IEEEbib}
\bibliography{strings,refs}

%\begin{thebibliography}{9}
%\end{thebibliography}

\end{document}